%% file: main.tex
\documentclass[10pt,twocolumn,letterpaper]{article}

\usepackage[pagenumbers]{cvpr} 
\usepackage{tabularx}
\usepackage{multirow}
\usepackage{listings}
\usepackage{tcolorbox}
\usepackage{cuted}
\newcolumntype{C}{>{\centering\arraybackslash}X}

\usepackage[accsupp]{axessibility}  


\definecolor{cvprblue}{rgb}{0.21,0.49,0.74}
\usepackage[pagebackref,breaklinks,colorlinks,allcolors=cvprblue]{hyperref}

\def\paperID{78}
\def\confName{EarthVision}
\def\confYear{2025}

\title{TerraMesh: A Planetary Mosaic of Multimodal Earth Observation Data}

\author{
    Benedikt Blumenstiel$^{1}$ \and
    Paolo Fraccaro$^{1}$ \and
    Valerio Marsocci$^{2}$ \and 
    Johannes Jakubik$^{1}$ \and
    Stefano Maurogiovanni$^{3,4}$ \and
    Mikolaj Czerkawski$^{2}$ \and
    Rocco Sedona$^{3}$ \and 
    Gabriele Cavallaro$^{3,4}$ \and
    Thomas Brunschwiler$^{1}$ \and  
    Juan Bernabe-Moreno$^{1}$ \and
    Nicolas Longépé$^{2}$ \vspace{2pt} \and
    $^{1}$IBM Research -- Europe\\
    $^{2}$European Space Agency $\Phi$-Lab\\
    $^{3}$Forschungszentrum J\"ulich\\
    $^{4}$University of Iceland\\
    {\tt\small benedikt.blumenstiel@ibm.com}
    \vspace{-2pt}
}

\begin{document}
\twocolumn[{
\maketitle
\vspace{-5mm}
\begin{center}
    \centering
    \includegraphics[width=\textwidth]{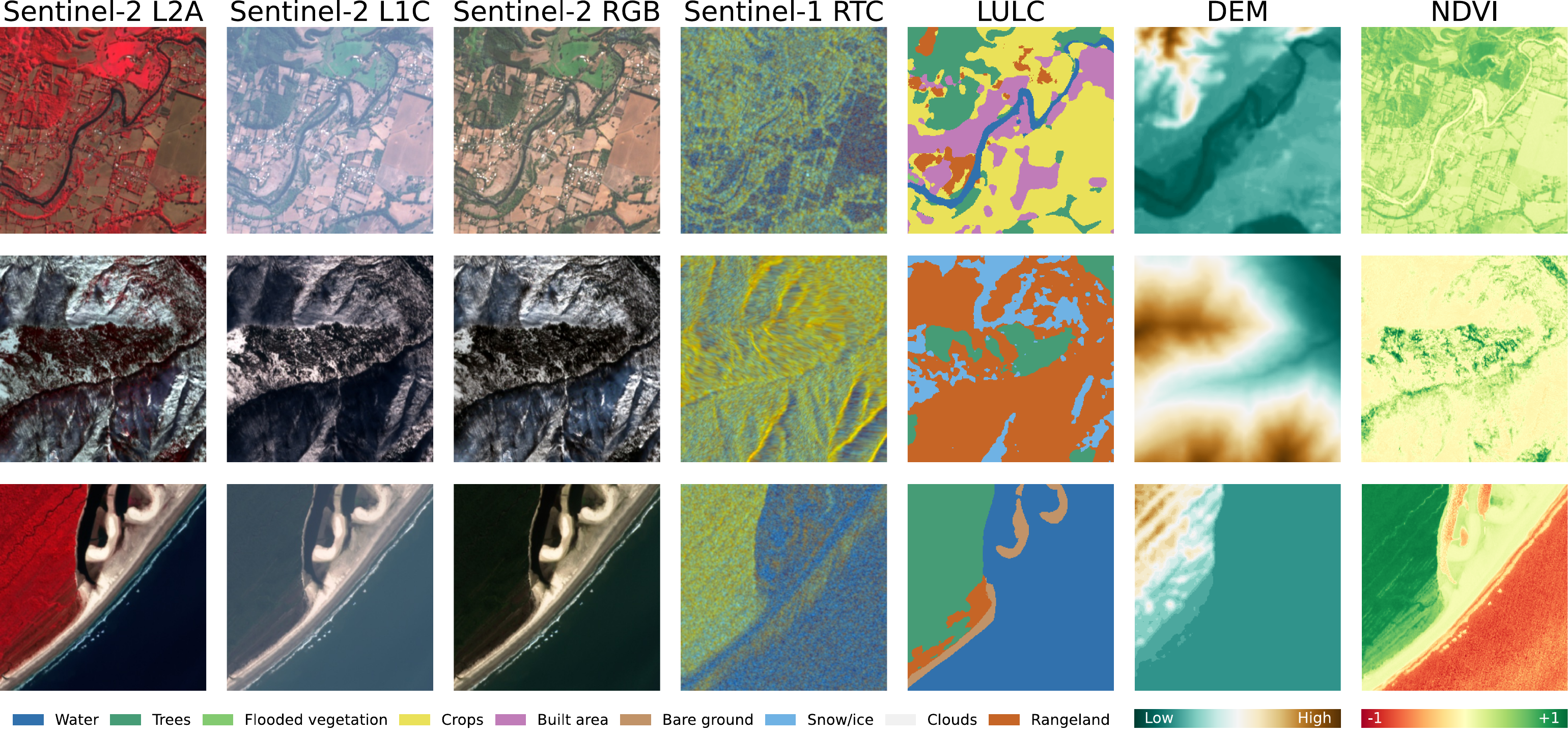}
    \captionof{figure}{Samples from the TerraMesh dataset with seven spatiotemporal aligned modalities. Sentinel-2 L2A uses IRRG pseudo-coloring, and Sentinel-1 RTC is visualized in dB scale as VH-VV-VV/VH. Copernicus DEM is scaled based on the image value range with an additional 10\,meter buffer to highlight flat scenes.}
    \label{fig:modalities}
\end{center}
}]

\input{content}
{
    \small
    \bibliographystyle{ieeenat_fullname}
    \bibliography{main}
}

\input{supplementary}

\end{document}

%% file: content.tex
\begin{abstract}
Large-scale foundation models in Earth Observation can learn versatile, label-efficient representations by leveraging massive amounts of unlabeled data. However, existing public datasets are often limited in scale, geographic coverage, or sensor variety. We introduce \textbf{TerraMesh}, a new globally diverse, multimodal dataset combining optical, synthetic aperture radar, elevation, and land-cover modalities in an Analysis-Ready Data format. TerraMesh includes over 9~million samples with eight spatiotemporal aligned modalities, enabling large-scale pre-training. 
We provide detailed data processing steps, comprehensive statistics, and empirical evidence demonstrating improved model performance when pre-trained on TerraMesh. The dataset is hosted at \url{https://huggingface.co/datasets/ibm-esa-geospatial/TerraMesh}.
\end{abstract}
\vspace{-10pt}

\begin{table*}[bth]
\centering
\caption{Comparison of publicly available EO pre-training datasets. The patch size refers to the Sentinel-2 samples. The sample count refers to single spatiotemporal samples with aligned modalities, and we report the length of the time series data.}
\label{tab:dataset_comparison}
\setlength\tabcolsep{2pt}
\begin{tabularx}{\textwidth}{lcCcccc}
    \toprule
    \textbf{Dataset} & \textbf{Year} & \textbf{Image modalities} & \textbf{Patch size} & \textbf{Sample count} & \textbf{Temporal} & \textbf{Coverage} \\
    \midrule
    So2Sat~\cite{so2sat2019} & 2019 & S-1, S-2 & 32$\times$32 & 400k & Single-season & Global cities \\
    SEN12MS~\cite{sen12ms2019} & 2019 & S-1, S-2, LULC & 256$\times$256 & 180k & Multi-season & Global \\
    BigEarthNet-MM~\cite{bigearthnetmm} & 2021 & S-1, S-2 & 120$\times$120 & 590k & Single-season & Europe \\
    SSL4EO-S12~\cite{ssl4eo2023} & 2023 & S-1, S-2 & 264$\times$264 & 250k $\times$ 4 & Multi-temporal & Global cities \\
    Satlas Pretrain~\cite{satlaspretrain2023} & 2023 & S-2, NAIP (partially) & 512$\times$512 & 856k $\times$ 8-12  & Multi-temporal & Global \\
    MajorTOM-Core~\cite{majortom2024} & 2024 & S-1, S-2, DEM & 1068$\times$1068 & 1.4M & Single-season & Global \\
    MMEarth~\cite{mmearth2024} & 2024 & S-1, S-2, DEM, GCHM, LULC & 128$\times$128 & 1.2M & Multi-season & Global \\ 
    \textbf{TerraMesh (ours)} & 2025 & S-1, S-2, DEM, NDVI, LULC & 264$\times$264 & 9M & Multi-season & Global \\ 
    \bottomrule
\end{tabularx}
\end{table*}

\let\thefootnote\relax\footnotetext{© 2024 IEEE. Personal use of this material is permitted. Permission from IEEE must be obtained for all other uses, in any current or future media, including reprinting/republishing this material for advertising or promotional purposes, creating new collective works, for resale or redistribution to servers or lists, or reuse of any copyrighted component of this work in other works.}

\section{Introduction}
\label{sec:intro}

Large-scale foundation models in Earth Observation (EO) promise universal, label-efficient feature learning across diverse landscapes and sensor types~\cite{prithvi2,galileo,anysat}. By capturing relationships within massive amounts of unlabeled data, these models can reduce the need for task-specific annotations, adapt to new regions or modalities more effectively, and ultimately provide stronger performance on downstream tasks related to land-cover mapping, change detection, and disaster response~\cite{prithvi2,galileo,anysat}. Despite these potential benefits, effective training of multimodal EO foundation models often remains limited by the lack of a large, well-curated dataset with consistently aligned sensor modalities. 

Given this limitation, new foundation models often require building proprietary pre-training sets~\cite{prithvi2,galileo,anysat,croma,clay}. A significant portion of these models focus on a single optical modality~\cite{prithvi2023,prithvi2,satlaspretrain2023,ssl4eo2023}, while others attempt to incorporate multiple sensors by processing modalities separately, not requiring spatiotemporal alignment~\cite{dofa,clay,anysat}. However, such approaches hinder the ability to learn meaningful cross-modal representations. To date, only a few approaches have leveraged fully co-registered datasets, which are limited to around one million samples~\cite{galileo,mmearth2024,decur}. 

We address this gap by introducing TerraMesh, a multimodal dataset specifically tailored for large-scale pre-training, which fuses optical, synthetic aperture radar (SAR), elevation, and other derived modalities into a single Analysis-Ready Data (ARD) resource. 
The dataset includes 9~million co-registered samples with 64~million individual image patches and has more than 20 times more pixel values than other public multimodal EO datasets~\cite{mmearth2024,ssl4eo2023}. 
Our dataset covers broad geographical extents, spans all seasons, and is released under a permissive CC-BY-SA-4.0 license. This enables correlation learning between modalities, which offers a more powerful pretext objective for models to learn meaningful embeddings~\cite{decur,galileo,anysat}.

In summary, our contribtions are: (1)~the largest publicly available co-registered multimodal EO dataset to date; (2) a thorough data filtering and subsampling pipeline ensuring diversity, quality, and ease of use due to the ARD format; and (3)~empirical evidence demonstrating that multimodal models pre-trained on TerraMesh outperform other state-of-the-art models on a variety of downstream tasks. This dataset will be valuable for advancing EO foundation models toward broader real-world impact.

\section{Related work}
\label{sec:related_work}

Developing large-scale, multimodal EO datasets has been critical for advancing self-supervised learning in remote sensing. As shown in Table~\ref{tab:dataset_comparison}, multiple works have curated training corpora that combine multiple sensors, typically focusing on Sentinel-1 (S-1) and Sentinel-2 (S-2) imagery due to their complementary nature, global coverage, and open accessibility. Existing datasets such as So2Sat~\cite{so2sat2019}, BigEarthNet~\cite{bigearthnet2019}, and SEN12MS~\cite{sen12ms2019} have provided large resources for classification tasks. Yet, these collections focus on a single timestamp and limited geographic and/or temporal sampling, which can restrict their suitability for pre-training foundation models that demand broader spatiotemporal variability. 

In contrast, newer efforts such as SSL4EO-S12~\cite{ssl4eo2023} and SatlasPretrain~\cite{satlaspretrain2023} have emphasized global-scale multi-sensor and multi-seasonal acquisitions. Empirical evidence shows that foundation models trained on these datasets can achieve strong performance in downstream tasks~\cite{decur,croma,satlaspretrain2023,ssl4eo2023}. In particular, experiments comparing SEN12MS, BigEarthNet, and SSL4EO-S12 as pre-training datasets show that scaling the dataset size leads to higher downstream task performance~\cite{ssl4eo2023}.
Meanwhile, MajorTOM-Core~\cite{majortom2024} addresses global geospatial coverage by including nearly all land surfaces. Yet, it comes with challenges to construct the ARD pre-training dataset due to the data format and unbalanced data distribution. 
Despite this progress, most large-scale datasets focus on a single pair of modalities (e.g., S-1 and S-2). MMEarth~\cite{mmearth2024} demonstrates a more diverse set of modalities with digital elevation models (DEM), vegetation annotations, and land-use/land-cover (LULC) products by leveraging pseudo-labeling. However, it is limited by a small patch size and overall sample count. In comparison, TerraMesh includes over 30 times more pixel values and is therefore better suited for generalization and model scaling due to the larger training corpus.

Beyond the public datasets, several proprietary datasets have been curated to pre-train EO foundation models. Galileo~\cite{galileo} uses 127k time series with nine co-registered modalities, while Prithvi~EO~2.0~\cite{prithvi2} leverages 4.2M samples of time series from Harmonized-Landsat-Sentinel (HLS) data. Other approaches combine multiple smaller datasets: msGFM~\cite{msgfm2024} is pre-trained with around 2M samples taken from SEN12MS and other RGB sources, and AnySat~\cite{anysat} merges five datasets covering nearly 2M locations and eleven different modalities. Besides their limited accessibility, many of these datasets remain constrained by their patch size and sample count, sensor diversity, spatiotemporal alignment, or limited global coverage.

TerraMesh addresses these issues by aligning optical, SAR, DEM, vegetation, and LULC data. Moreover, it is globally distributed, with urban areas specifically upsampled, and has temporally diverse data. The dataset is designed to be pre-training ready and is publicly available.

\begin{table*}[bth]
    \centering
    \caption{Summary of modalities included in TerraMesh. The range includes the typical value range. Every co-registered sample includes seven out of eight modalities due to the source datasets providing either S-1 GRD or S-1 RTC data.}
    \label{tab:modalities}
    \begin{tabularx}{\textwidth}{lccccX}
        \hline
         \textbf{Sensor/Product} & \textbf{Type} & \textbf{Scale} & \textbf{Range} & \textbf{\#Bands} & \textbf{Band information} \\
        \hline
        Sentinel-2 L1C & Optical & DN & 0 -- 10000 & 13 & Top-of-atmosphere reflectance \\
        Sentinel-2 L2A & Optical & DN & 0 -- 10000 & 12 & Bottom-of-atmosphere reflectance \\
        Sentinel-2 RGB & Optical & RGB & 0 -- 255 & 3 & Processed images based on S-2 L2A \\
        Sentinel-1 GRD & SAR & dB & -50 -- +10 & 2 & VV, VH, calibrated SAR backscatter \\
        Sentinel-1 RTC & SAR & dB & -50 -- +10 & 2 & VV, VH, radiometrically terrain-corrected \\
        NDVI & Index & -- & -1 -- +1 & 1 & Normalized Difference Vegetation Index \\
        Copernicus DEM & Elevation & Meter & -400 -- 8800 & 1 & Digial Elevation Model based on TanDEM-X \\
        LULC & Annotations & Classes & 0 -- 9 & 1 & Land-use/land-cover labels from ESRI \\
        \hline
    \end{tabularx}
\end{table*}

\section{Dataset}
\label{sec:dataset}

This section provides an overview of the included modalities and the applied processing steps. This includes the selection of data sources and subsampling, followed by the data processing pipeline. Lastly, we describe the final dataset statistics and the data format.

\subsection{Modalities}
\label{subsec:modalities}

All modalities in TerraMesh are co-registered to a consistent grid, based on Sentinel-2 with a 10~m resolution, ensuring precise spatial alignment across the different sensors. Acquisitions are temporally matched, with the majority of the S-1 and S-2 scenes collected within a one-week window. 

Table~\ref{tab:modalities} provides an overview of each modality in TerraMesh, including its value range, band information, and general product characteristics.
TerraMesh includes multiple optical data sources from Sentinel-2. Level-1C (L1C) offers top-of-atmosphere reflectances, while Level-2A (L2A) provids bottom-of-atmosphere reflectance values. 
Providing L1C and L2A ensures that pre-trained models can be directly used for many downstream tasks regardless of the processing level.
We also generate an RGB product (S-2 RGB) from L2A to address downstream tasks relying on RGB imagery. Following literature~\cite{majortom2024,ssl4eos12v11}, we additionally provide SEnSeI v2 cloud masks with the optical data.
Next, we incorporate two synthetic aperture radar (SAR) products from Sentinel-1. Ground Range Detected (GRD) data undergoes thermal noise removal and calibration, while Radiometrically Terrain Corrected (RTC) data further compensates for terrain effects and ensures precise spatial alignment. Due to different availability in the source datasets, every sample includes only one S-1 processing level.

An NDVI layer derived from S-2 L2A complements these satellite observations and serves as a robust vegetation index and suitable pretext task for agricultural and forestry-related downstream applications.
TerraMesh also leverages the Copernicus DEM, a global 30~m resolution elevation model based on TanDEM-X, to provide topographic context. For land-cover monitoring, we include a yearly LULC product from ESRI, augmented with SEnSeI-based annotations to better capture temporally variable classes such as clouds and ice. 

We explored additional modalities via pseudo-labeling using the same AI models as MMEarth (LULC via DynamicWorld~\cite{dynamicworld} and canopy tree height maps via ETH-GCHM~\cite{treeheight}). However, we observed limited generalization with many false predictions, e.g., \textit{ice/snow} class predictions in the desert or many negative tree height values.
We assume this is due to limited temporal coverage in the model training data, which does not generalize to other years or data sources. This might be due to atmospheric variances between seasons and years, and/or different processing versions for Sentinel-2 data. Specifically, DynamicWorld was trained on S-2 data from 2019~\cite{dynamicworld}, and ETH-GCHM uses S-2 data from 2020~\cite{treeheight}. The S-2 processing pipeline changed significantly in 2021, including geometric refinement and improved radiometric harmonization\footnote{S-2 processing versions: \url{https://sentiwiki.copernicus.eu/web/s2-processing}}. We compared the original training data with updated data using the newer processing version and observed large variances in the reflectance values that might explain the false predictions. Adding a fixed offset or scaling could not reduce the false predictions. In comparison, SEnSeI v2~\cite{senseiv2} generalized well to images from all years and data sources, potentially due to its training dataset being a diverse collection of multiple cloud datasets. 

\subsection{Data Processing}
\label{subsec:processing}

Our dataset objective is a large-scale, multimodal corpus with broad spatiotemporal coverage. To achieve this, TerraMesh merges SSL4EO-S12\,v1.1~\cite{ssl4eos12v11} with a subset of MajorTOM-Core~\cite{majortom2024}, thereby combining near-global coverage from MajorTOM-Core with multi-seasonal imagery of urban areas from SSL4EO-S12. Both datasets include S-2 L1C, S-2 L2A, and S-1 GRD/RTC products, which simplifies integration. However, they only include one S-1 product each, and we process the data separately to ensure spatiotemporal alignment between seven out of eight modalities.
MajorTOM-Core covers nearly every location on land, resulting in numerous highly homogeneous samples (e.g., large deserts). Previous research has shown that this can destabilize model training due to very low reflectance values in water patches or high values over ice or sand~\cite{prithvi2}. Below, we describe how we subsample MajorTOM-Core to improve data diversity and then detail the steps to harmonize the samples into an ARD standard. We conclude by outlining the processing steps for the extended set of modalities.

\begin{figure*}[tbh]
    \centering
    \includegraphics[width=\linewidth]{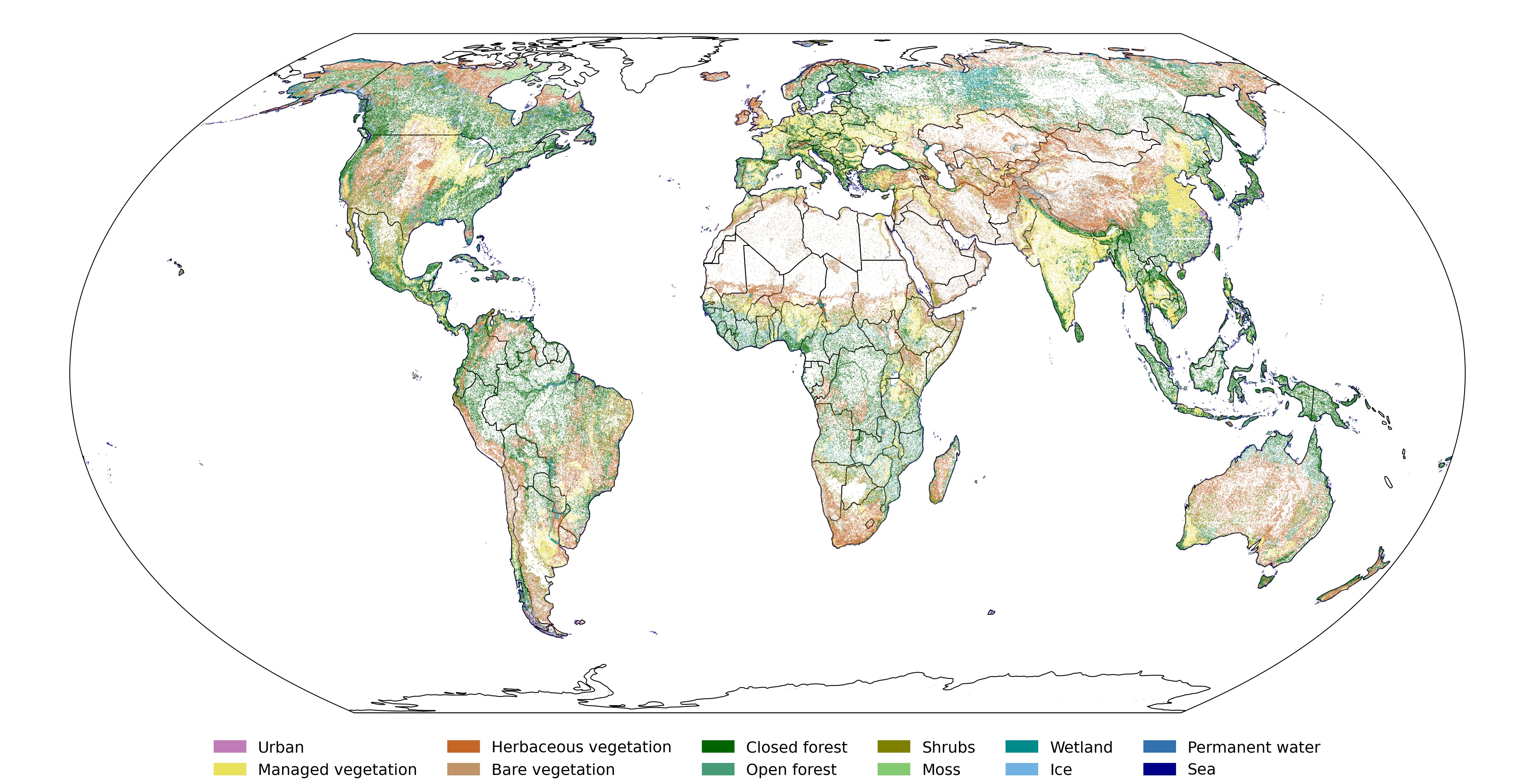}
    \caption{All selected grid cells from MajorTOM-Core visualized with their majority class from Copernicus Land Cover 100m~\cite{landcover2019}.}
    \label{fig:map}
\end{figure*}

\paragraph{Subsampling}
Following the approach in~\cite{prithvi2}, we aim to create a diverse and representative dataset covering all landcover types and worldwide ecoregions. We begin by discarding any MajorTOM-Core grid cells containing more than 50\% missing data (\texttt{NaN}) and retain only those where Sentinel-1 and Sentinel-2 acquisitions overlap, selecting the pairs with minimal time difference. We then compute the class coverage of each S-2 grid cell with Copernicus Land Cover 100m~\cite{landcover2019} and RESOLVE Ecoregions~\cite{ecoregions2017}. The original 12 Copernicus forest categories are merged into two aggregated \textit{closed forest} and \textit{open forest} classes similar to~\cite{prithvi2}. For each land-cover category, we select grid cell candidates within 5\%--99\% coverage to avoid extremely homogeneous patches and sample up to 500k grid cells. Subsequently, we ensure that each of the 846 ecoregions is represented by at least 500 grid cells, provided sufficient candidate patches exist in that region with at least 5\% coverage. The resulting selection comprises 538k grid cells, depicted in Figure~\ref{fig:map}. Large homogeneous regions like the Sahara are noticeably downsampled. Some regions are not well covered due to missing S-1 samples in the source datasets, particularly Antarctica, Greenland, and a few other islands.

\paragraph{Preprocessing}
In assembling TerraMesh, we leveraged data from multiple repositories. MajorTOM-Core provides S-2 data from the CREODIAS platform and S-1 data from Planetary Computer~\cite{majortom2024}, whereas SSL4EO-S12\,v1.1 relies on Google Earth Engine for its S-1 and S-2 acquisitions~\cite{ssl4eos12v11}. We obtained the Copernicus DEM from the Copernicus Data Space Ecosystem\footnote{Copernicus DEM source: \url{https://dataspace.copernicus.eu/explore-data/data-collections/copernicus-contributing-missions/collections-description/COP-DEM}}, and further acquired ESRI’s LULC maps from Planetary Computer\footnote{ESRI LULC source: \url{https://planetarycomputer.microsoft.com/dataset/io-lulc-annual-v02}}.

We merge the selected MajorTOM-Core grid cells with the SSL4EO-S12\,v1.1 dataset. Each grid cell is a 1068\,$\times$\,1068-pixel tile covering 16 non-overlapping patches at the 264\,$\times$\,264 pixel size of SSL4EO-S12~\cite{ssl4eo2023}. 
This patch size, divisible by six, accommodates Sentinel-2’s resolutions of 10\,m, 20\,m, and 60\,m. Prior to patch generation, we reproject Sentinel-1 acquisitions to the Sentinel-2 grid in cases of differing UTM zones. 

A random 99\%-1\% train-validation split is applied to the selected MajorTOM-Core grid cells. Any SSL4EO-S12\,v1.1 patch that intersects the validation extent is added to the validation set; further patches spatially overlapping the joined validation area are removed to mitigate data leakage. All Sentinel-2 bands are resampled to 10\,m using nearest interpolation, and the +1000 offset is removed from post-2022 data, which was added in ESA’s latest processing standard~\cite{s2offset}.

Next, we discard any patch with more than 1\% missing values (\texttt{NaN}) in any modality or channel, and fill the remaining \texttt{NaN}s via nearest-neighbor interpolation. Because the provided cloud masks from Sentinel-2 can be unreliable, we follow \cite{majortom2024,ssl4eos12v11} and employ the SEnSeI v2 model~\cite{senseiv2} to generate accurate cloud and ice masks.

For the RGB visualization, we produce Sentinel-2 RGB composites from L2A products using the pseudocode illustrated in Listing~\ref{lst:rgb} in the supplementary material, which ensures a balanced color representation. To avoid saturation in bright regions (e.g., snow or sand), we do not use a fixed reflectance range but perform quantile-based adjustments. First, extreme values below the 2\% and above the 98\% quantile are reduced using a factor of 0.5. Using the 0.2\% and 99.8\% quantiles as the lower and upper limits, we then map the adjusted values to an 8-bit (0--255) range. If the upper limit falls below 2000, we clamp it to 2000. If the median of the RGB values is below 1000, we set the lower limit to 0 to prevent further darkening already dim images. 

We also provide NDVI layers derived from Sentinel-2 L2A, by clipping negative values to zero and adding a small epsilon in the denominator. Copernicus DEM data, acquired at 30\,m resolution, is bilinearly resampled and projected to the Sentinel-2 grid. ESRI LULC products at 10\,m resolution are downloaded for the closest available year from the S-2 timestamp and regridded correspondingly. Missing LULC pixels are assigned the \textit{no data} class. To improve the alignment between the satellite images and the LULC maps, we replace the ESRI \textit{cloud} class with \textit{no data} and add the \textit{cloud} and \textit{snow/ice} predictions from the SEnSeI~v2 masks. 

\paragraph{Storage format}
We finalize the dataset by saving each patch as a Zarr Zip file (\texttt{ZipStore}, version~2\footnote{Zarr documentation: \url{https://zarr-specs.readthedocs.io/en/latest/v2/v2.0.html}}) using \texttt{xarray Datasets}. 
The patches are shuffled and stored in tar archives that can be used with WebDataset\footnote{WebDataset documentation: \url{https://github.com/webdataset/webdataset}}.
This approach consolidates large numbers of samples into fewer individual files, which is easier to handle on HPC clusters and other file systems with a limited number of inodes. 
We further optimize the data size by storing the Sentinel-2 bands and DEM as 16-bit integers, while Sentinel-1 and NDVI are saved as 16-bit floats; both LULC and Sentinel-2 RGB are stored in 8-bit format. In conjunction with Zarr’s built-in compression, these measures substantially reduce storage overhead without sacrificing data fidelity. 
When comparing a TIF data format and Zarr, the dataset size reduces from 41\,TB to only 16\,TB, achieving a compression factor of 2.9. The compression factor varies between modalities, ranging from 20 for LULC to only 2.5 for S-2 RGB. By total size, S-2 compression has the biggest effect, saving 18.6\,TB with factors of 2.6 (S-2 L2A) and 2.9 (S-2 L1C). 
Using WebDataset with Zarr Zip files also optimizes the data loading speed, as demonstrated in the supplementary material.

\subsection{Dataset description}
\label{subsec:description}

\begin{figure}[tbh]
    \centering
    \includegraphics[width=\linewidth]{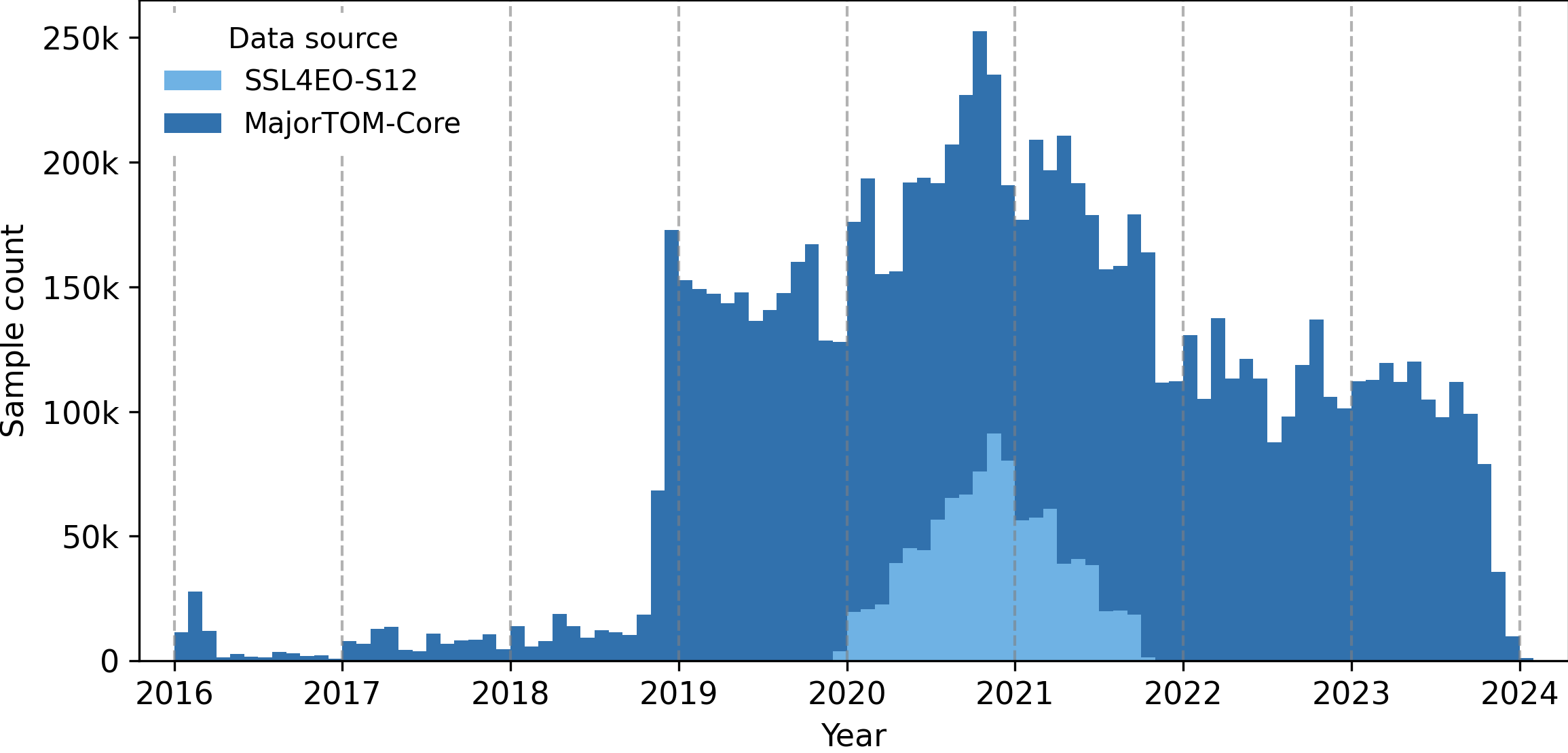}
    \caption{Monthly distribution of all S-2 timestamps.}
    \label{fig:temporal}
\end{figure}

TerraMesh includes 16\,TB of compressed data and consists of 9\,089\,536 training and 89\,088 validation samples, aggregated into tar archives per modality for streamlined data access with WebDataset. Each spatiotemporal sample features seven aligned modalities, resulting in 64M individual patches, all modalities considered. Figure~\ref{fig:temporal} illustrates the temporal distribution of these samples, using the S-2 timestamp. The data covers eight years with the majority of acquisitions between 2019 and 2023.

\begin{figure}[bth]
    \centering
    \includegraphics[width=0.8\linewidth]{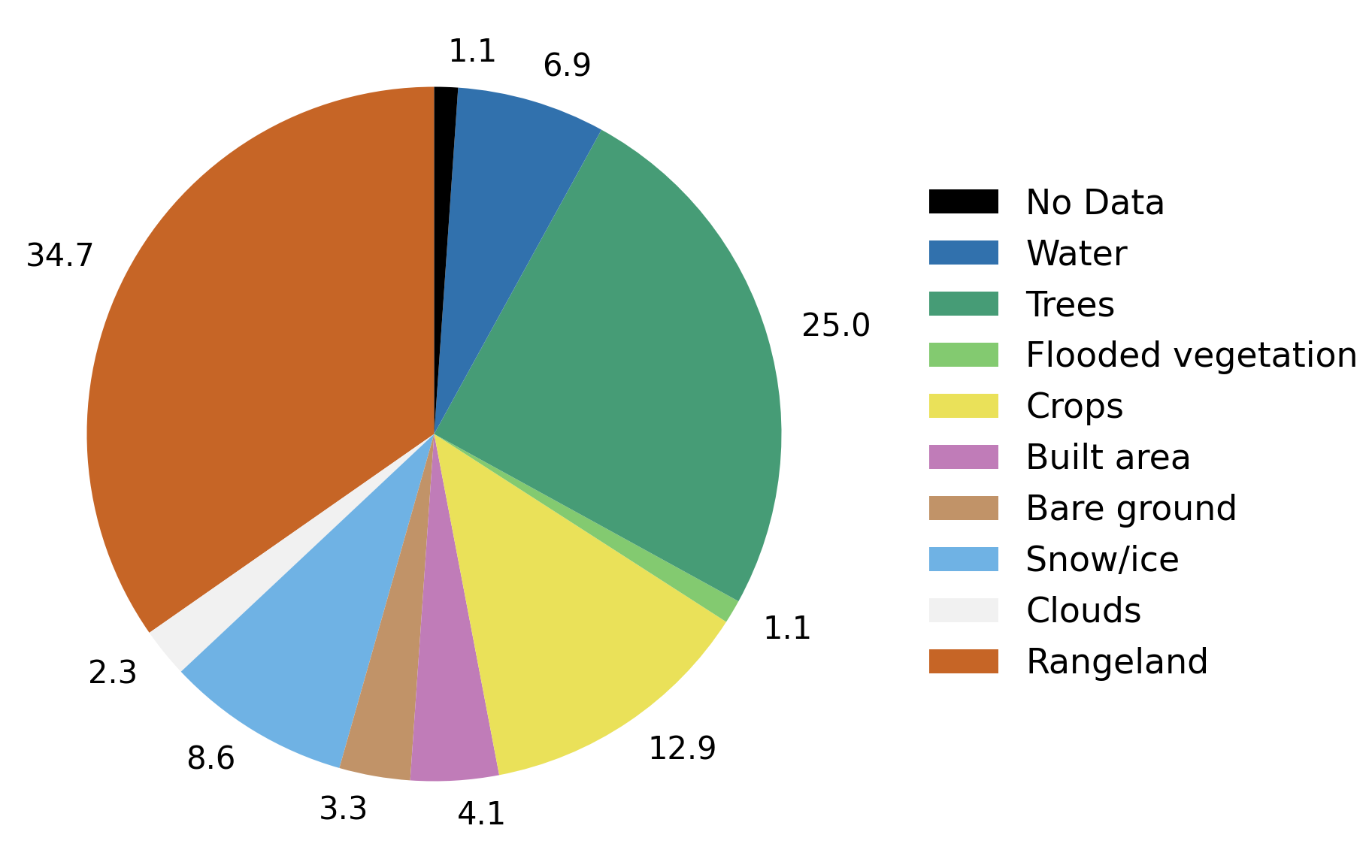}
    \caption{ESRI LULC class distribution in TerraMesh.}
    \label{fig:lulc}
\end{figure}

\begin{figure}[tbh]
    \centering
    \includegraphics[width=\linewidth]{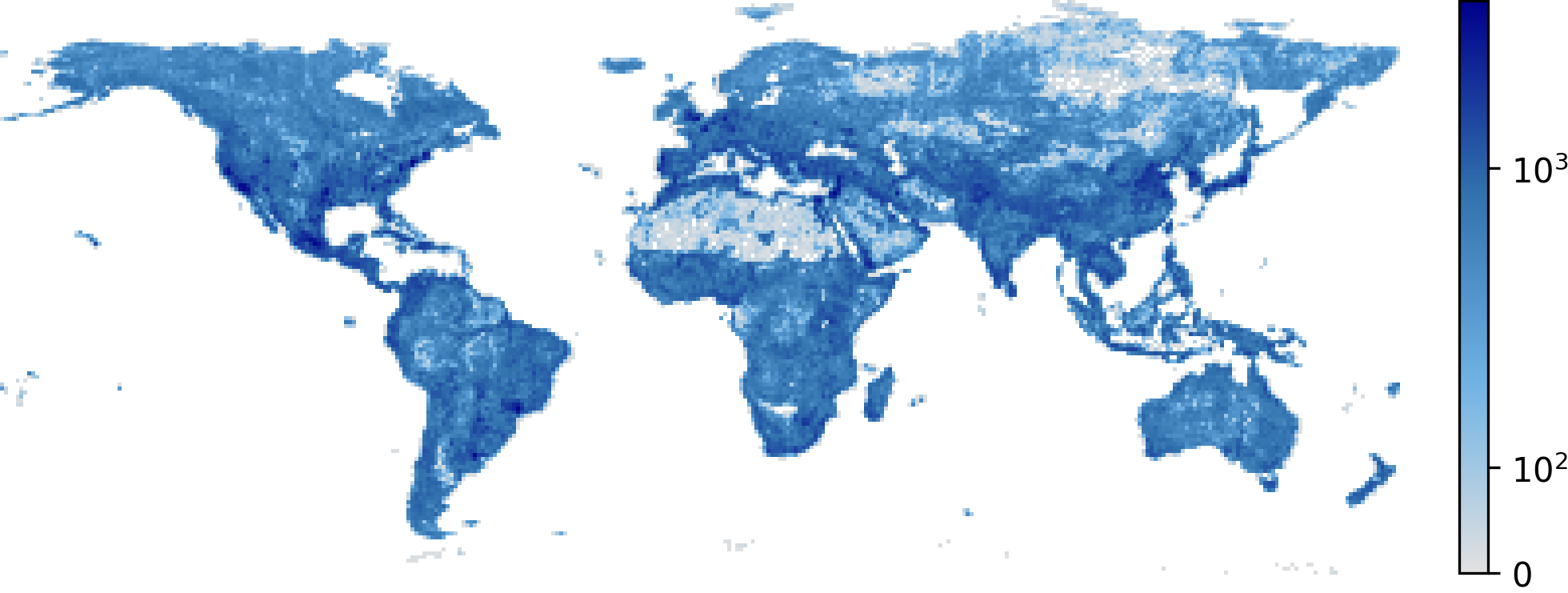}
    \caption{Heat map of the sample count in a one-degree grid.}
    \label{fig:sample_count}
\end{figure}

Figure~\ref{fig:lulc} details the distribution of ESRI LULC classes within the dataset. Subsampling MajorTOM-Core and adding the data from SSL4EO-S12, which is urban-centered, increased the build area to 4\%, compared to less than 1\% of urban areas on all land masses~\cite{landcover2019,prithvi2}. Furthermore, we see a high proportion of 34.7\% rangeland and 25.0\% forests in TerraMesh compared to only 28.6\% and 14.2\% in SSL4EO-S12. The urban-centered dataset also has twice the ratios for crop and built-up areas. In the supplementary material, we provide visual examples from the source datasets to display the different types of covered landscapes in both datasets. 

Figure~\ref{fig:sample_count} shows the global distribution of samples using a log scale. Unlike smaller datasets, like So2Sat or BigEarthNet, it fully includes the southern hemisphere and at least a few examples from nearly all geographies, like Antarctica or remote islands. Homogeneous regions like the Sahara or tundra are downsampled by design.


All essential metadata is colocated with the image data in the Zarr archives, including spatial references (latitude/longitude in EPSG:4326 and per-pixel coordinates in the corresponding UTM zones), acquisition timestamps, and the SEnSeI~v2 cloud masks. These masks distinguish among \emph{land}, \emph{water}, \emph{snow}, \emph{thin cloud}, \emph{thick cloud}, and \emph{cloud shadow}. Embedding the metadata and masks alongside the imagery simplifies data management and enables users to extract information about the images seamlessly.

TerraMesh is made publicly available under the CC-BY-SA-4.0 license, permitting broad adoption and reuse in both academic research and industrial applications.

\section{Experiments}
\label{sec:experiments}

\begin{table*}[tbh]
\centering
\setlength{\tabcolsep}{3pt}
\caption{Performance evaluation of TerraMind across six benchmark datasets using the PANGAEA evaluation protocol. The mIoU $\uparrow$ is reported. We summarize the results with average mIoU and average rank $\downarrow$. The two best results are highlighted in bold and underlined.}
\label{tab:results}
\begin{tabularx}{\textwidth}{lCCCCCCcc}
\toprule
Model & BurnSr & MADOS & PASTIS & Sen1Fl11 & CTM-SS & SN7 & Avg. mIoU & Avg. Rank \\
\midrule
CROMA                & 82.42 & \underline{67.55} & 32.32 & \textbf{90.89} & 49.38 & 59.28 & 63.64 & 4.50 \\
DOFA                 & 80.63 & 59.58 & 30.02 & 89.37 & 51.33 & 61.84 & 62.13 & 6.17 \\
GFM-Swin             & 76.90 & 64.71 & 21.24 & 72.60 & 46.98 & 60.89 & 57.22 & 8.83 \\
Prithvi 1.0 100M     & \underline{83.62} & 49.98 & 33.93 & 90.37 & 43.07 & 56.54 & 59.59 & 8.17 \\
RemoteCLIP           & 76.59 & 60.00 & 18.23 & 74.26 & 52.05 & 57.76 & 56.48 & 9.33 \\
SatlasNet            & 79.96 & 55.86 & 17.51 & 90.30 & 46.97 & \underline{61.88} & 58.75 & 8.17 \\
Scale-MAE            & 76.68 & 57.32 & 24.55 & 74.13 & 25.42 & \textbf{62.96} & 53.51 & 9.67 \\
SpectralGPT          & 80.47 & 57.99 & 35.44 & 89.07 & 46.95 & 58.86 & 61.46 & 7.67 \\
SSL4EO-S12-MoCo             & 81.58 & 51.76 & 34.49 & 89.26 & 48.58 & 57.64 & 60.55 & 7.83 \\
SSL4EO-S12-DINO             & 81.72 & 49.37 & 36.18 & 88.61 & 48.66 & 56.47 & 60.17 & 8.50 \\
SSL4EO-S12-MAE              & 81.91 & 49.90 & 32.03 & 87.79 & 45.80 & 57.13 & 59.09 & 10.00 \\
SSL4EO-S12-Data2Vec         & 81.91 & 44.36 & 34.32 & 88.15 & \textit{54.03} & 58.23 & 60.17 & 7.50 \\
\midrule
\textbf{TerraMind-B (TerraMesh)}  & 82.42 & \textbf{69.52} & \underline{40.51} & \underline{90.62} & \textbf{55.80} & 60.61 & \textbf{66.58} & \textbf{2.33} \\
\midrule
\textbf{TerraMind-B (Ablation)}   & \textbf{84.00} & 65.01 & \textbf{40.80} & 90.32 & 52.66 & 59.71 & \underline{65.42} & \underline{3.00} \\
\bottomrule
\end{tabularx}
\end{table*}

The dataset’s utility is demonstrated using TerraMind~\cite{terramind}, a  4M-like model~\cite{4m} pre-trained on TerraMesh. 
The model is pre-trained in a two-stage framework: first, each modality is encoded via VQ-VAE tokenization; then, a correlation-learning approach employs masked token prediction to align and fuse multimodal features~\cite{4m,terramind}. Similar to the raw RGB input modality in 4M, TerraMind uses the S-1 and S-2 satellite data as raw inputs. We also train a second model as an ablation to showcase the effect of scaling modalities and dataset size. For the ablation, we only use the SSL4EO-S12\,v1.1 locations with fewer modalities (S-2 L2A, S-1 GRD, DEM, and LULC). Both models use a ViT-B/16~\cite{vit} as the encoder and are pre-trained for 500B tokens. I.e., the ablation model is pre-trained with the same compute but with less diverse training data.

We benchmark TerraMind on six tasks from PANGAEA~\cite{pangaea} in Table \ref{tab:results}. Specifically, we selected five tasks that can be considered in-domain of TerraMesh, being semantic segmentation with Sentinel-2 or HLS data. However, we ensured to have a strong diversity in the application domain, ranging from agricultural (PASTIS~\cite{pastis}) to flood detection (Sen1Floods11~\cite{sen1floods11}) to a marine task (MADOS~\cite{mados}). Moreover, we added another dataset (i.e., SpaceNet7~\cite{spacenet7}), totally out-of-domain, both for the task (change detection) and input data (Maxar).
All models are fine-tuned with frozen encoders following the PANGAEA setting, which is a good evaluation for testing and comparing the model embeddings~\cite{pangaea}.

TerraMind, pre-trained on TerraMesh, outperforms most models on four datasets and is on par with other models in the remaining two tasks. The model reaches the best average rank and an average mIoU of 66.58\%, which is 3\,pp. higher than the next-best model architecture CROMA~\cite{croma}. This demonstrates the overall benefits of pre-training multimodal models on a large-scale dataset.

The TerraMind ablation is performing overall second-best, outperforming other models like CROMA and all SSL4EO-S12 models, which are pre-trained on the same image locations. This demonstrates the general capabilities of the TerraMind approach with correlation learning.

The ablation reaches the best result for the datasets BurnScars~\cite{burnscars} and PASTIS, outperforming the TerraMesh version. The datasets are sourced from the US and France, potentially benefiting from a more similar data distribution as both countries are well covered in SSL4EO-S12. In comparison, the larger dataset improves performance for crop type mapping in South Sudan (CTM-SS~\cite{ctmss}), maritime debris detection (MADOS), and the out-of-domain task SpaceNet 7 (SN7), which benefit from the increased data diversity due to more locations and modalities.

Overall, the observations are supported by other literature, which shows that scaling the pre-training data size improves the downstream task performance and is beneficial for model scaling~\cite{ssl4eo2023,prithvi2}. Furthermore, ablation experiments by \cite{galileo} and \cite{croma} demonstrate that additional pre-training modalities help in most downstream use cases.


\section{Conclusion}
\label{sec:conclusion}

This paper introduces TerraMesh, a large-scale dataset curated for multimodal Earth Observation pre-training. By combining and harmonizing existing resources, we offer a comprehensive corpus that aligns multiple EO modalities—including optical, SAR, elevation, and land-cover annotations–with a globally diverse spatiotemporal coverage. Our thorough subsampling and preprocessing ensure both data diversity and high quality, while Zarr-based packaging facilitates efficient large-scale training workflows.  

Experiments with TerraMind demonstrate the benefits of pre-training on TerraMesh, achieving state-of-the-art performance across various tasks. We believe TerraMesh opens new frontiers in EO research, empowering a broad community to develop new foundation models that fuse diverse remote sensing modalities in a spatially and temporally coherent manner.

\section{Acknowledgement}

This work is part of the Fostering Advancements in Foundation Models via Unsupervised and Self-supervised Learning for Downstream Tasks in Earth Observation (FAST-EO) project, which is funded by the European Space Agency (ESA) $\Phi$-Lab under contract No. 4000143501/23/I-DT.

%% file: supplementary.tex
\clearpage
\setcounter{page}{1}
\maketitlesupplementary

This supplementary material includes 
(1)~pseudocode for generating the additonal Sentinel-2 modalities,
(2)~an evaluation of different data formats,
and (3)~visual examples.

We provide pseudocode for generating Sentinel-2 RGB images and NDVI maps from Sentinel-2 L2A in Listing~\ref{lst:rgb} and~\ref{lst:ndvi}. The RGB algorithm first scales very dark and bright pixels before mapping the image to a uint8 range. We refer to Subsection~\ref{subsec:processing} for a detailed description.

\vspace{-5pt}

\begin{figure}[bh]
\begin{lstlisting}[label={lst:rgb}, caption=Pseudocode for the S-2-to-RGB transformation.]
\end{lstlisting}
\begin{tcolorbox}[boxrule=0pt, sharp corners, width=0.98\linewidth, boxsep=3pt, left=5pt, right=3pt, top=0pt, bottom=0pt]
\begin{lstlisting}[language=Python]
import numpy as np

def transform_rgb(rbg):
    Q2, Q98 = np.quantile(rbg, [0.02, 0.98])
    rbg = np.where(rbg >= Q2, rbg, 
                   Q2 + (rbg - Q2) * 0.5)
    rbg = np.where(rbg <= Q98, rbg, 
                   Q98 + (rbg - Q98) * 0.5)    
    Q02, Q50, Q998 = np.quantile(rbg, 
        [0.002, 0.5, 0.998])
    U = max(2000, Q998)
    L = 0 if Q50 < 1000 else Q02
    rbg = (rbg - L) / (U - L) * 255
    
    rbg = np.clip(rbg, 0, 255)
    return rbg.astype(np.uint8)

\end{lstlisting}
\end{tcolorbox}
\end{figure}

\vspace{-15pt}

\begin{figure}[bth]
\begin{lstlisting}[label={lst:ndvi}, caption=Pseudocode for the S-2-to-NDVI calculation.]
\end{lstlisting}
\begin{tcolorbox}[boxrule=0pt, sharp corners, width=0.98\linewidth, boxsep=3pt, left=5pt, right=3pt, top=0pt, bottom=0pt]
\begin{lstlisting}[language=Python]
def calculate_ndvi(b04, b08):
    b04 = (b04).clip(0)
    b08 = (b08).clip(0)

    ndvi = (b08 - b04) / (b08 + b04 + 1e-6)

    return ndvi
\end{lstlisting}
\end{tcolorbox}
\end{figure}

\vspace{-5pt}

We compare different dataset formats in terms of data loading speed and storage requirements in Table~\ref{tab:loading_speed}, using the validation split. The experiments were performed on an OpenShift cluster with an average expected I/O speed of 12.5GB/s between data storage and compute nodes.
The data loading speed is averaged over 1000 batches using eight CPUs and workers, excluding the initial batches due to slower processing times. The data size is extrapolated based on the size of the validation split.

We evaluate the performance of WebDataset, with individual samples stored as Zarr Zip, Numpy, or TIF files in tar archives. Additionally, we test Zarr Zip files containing batches of 64 samples as utilized by SSL4EO-S12~v1.1~\cite{ssl4eos12v11}. Zarr Zip files prove to be the most efficient for storage space, while WebDataset with Numpy files offers the fastest I/O speeds with eigth workers (3.6~GB/s). However, Numpy and TIF formats result in substantial storage needs, requiring up to 41.3 TB for the entire dataset. Combining WebDataset with Zarr Zip files optimizes storage space and data loading speed. With 16 workers, WebDataset with Zarr Zip reaches a speed of 0.047~s/batch outperforming Numpy files with 0.054~s/batch as I/O becomes more important than CPU processing. However, further adding CPUs leads to limited improvement.
Notably, Hugging Face restricts hosting large-scale datasets to WebDataset or parquet formats.
Further experiments with parquet files demonstrated strictly slower data loading times compared to WebDataset.

\begin{table}[bth]
\centering
\caption{Comparison of data loading speed averaged over 1000 batches with a batch size of 64. WebDataset with Zarr Zip files per sample best optimizes loading speed and storage requirements.}
\label{tab:loading_speed}
\setlength\tabcolsep{2pt}
\begin{tabularx}{\linewidth}{lXcc}
    \toprule
    \textbf{Mod.} & \textbf{Format} & \textbf{Speed} & \textbf{Data size} \\
    \midrule
    \multirow[c]{4}{0.5in}{All} & WebDataset w/ Zarr & 0.096 s/batch & 15.4 TB \\
    & WebDataset w/ Npy & 0.073 s/batch & 37.9 TB \\
    & WebDataset w/ TIF & 0.172 s/batch & 41.3 TB \\
    & Zarr Zip & 0.122 s/batch & 14.2 TB \\
    \midrule
    \multirow[c]{4}{0.5in}{S-2 L2A} & WebDataset w/ Zarr & 0.033 s/batch & 5.5 TB \\
    & WebDataset w/ Npy & 0.022 s/batch & 14.0 TB \\
    & WebDataset w/ TIF & 0.047 s/batch & 14.0 TB \\
    & Zarr Zip & 0.026 s/batch & 5.3 TB \\
    \midrule
    \multirow[c]{4}{0.5in}{S-1 RTC} & WebDataset w/ Zarr & 0.007 s/batch & 1.5 TB \\
    & WebDataset w/ Npy & 0.005 s/batch & 2.2 TB \\
    & WebDataset w/ TIF & 0.019 s/batch & 4.2 TB \\
    & Zarr Zip & 0.009 s/batch & 1.4 TB \\
    \bottomrule
\end{tabularx}
\end{table}

\begin{figure*}[tbh]
    \centering
    \includegraphics[width=\textwidth]{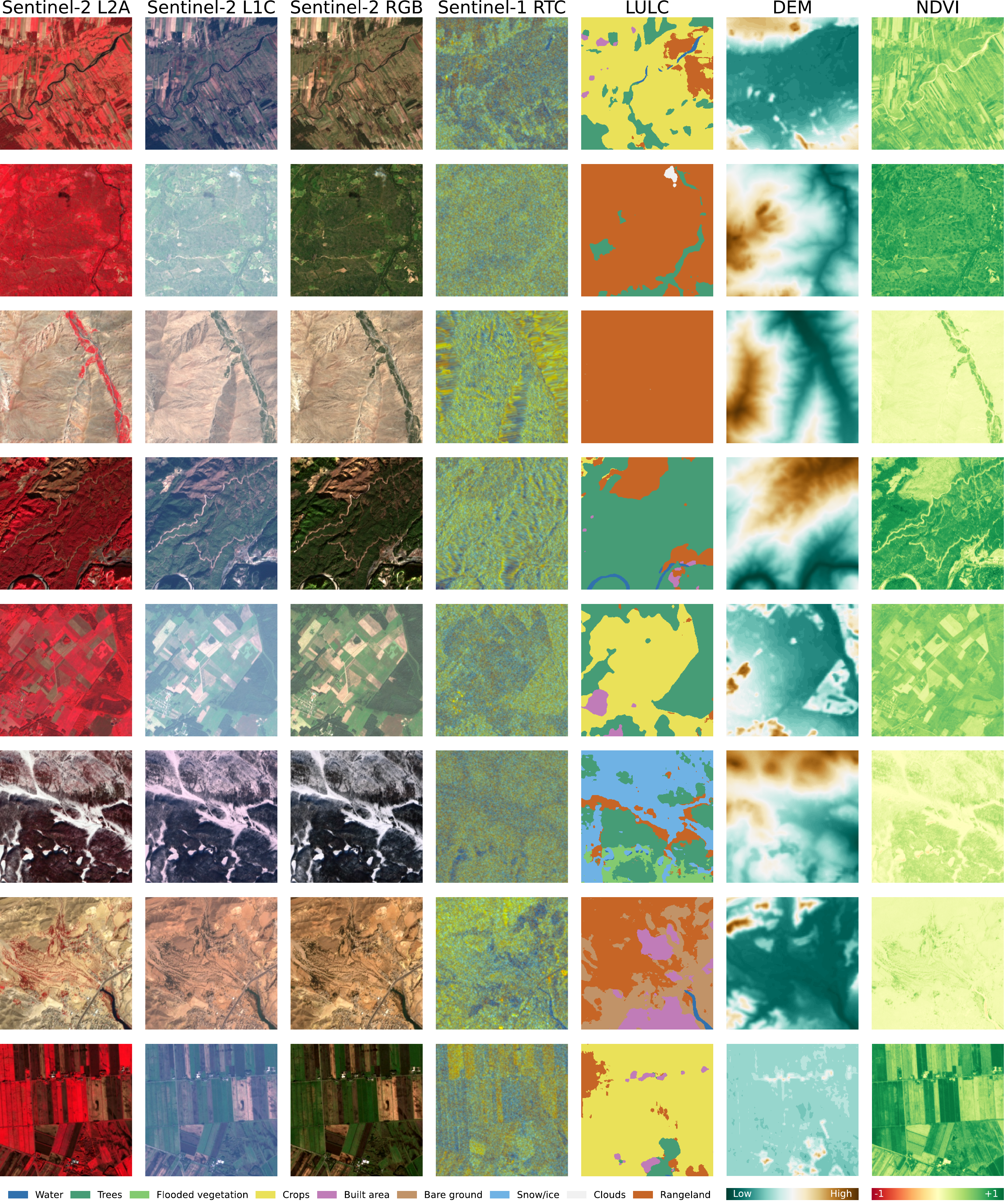}
    \captionof{figure}{Additional examples from the MajorTOM-Core subset with seven spatiotemporal aligned modalities. Sentinel-2 L2A uses IRRG pseudo-coloring, and Sentinel-1 RTC is visualized in dB scale as VH-VV-VV/VH. Copernicus DEM is scaled based on the image value range with an additional 10\,meter buffer.}
    \label{fig:majortom_samples}
\end{figure*}

\begin{figure*}[tbh]
    \centering
    \includegraphics[width=\textwidth]{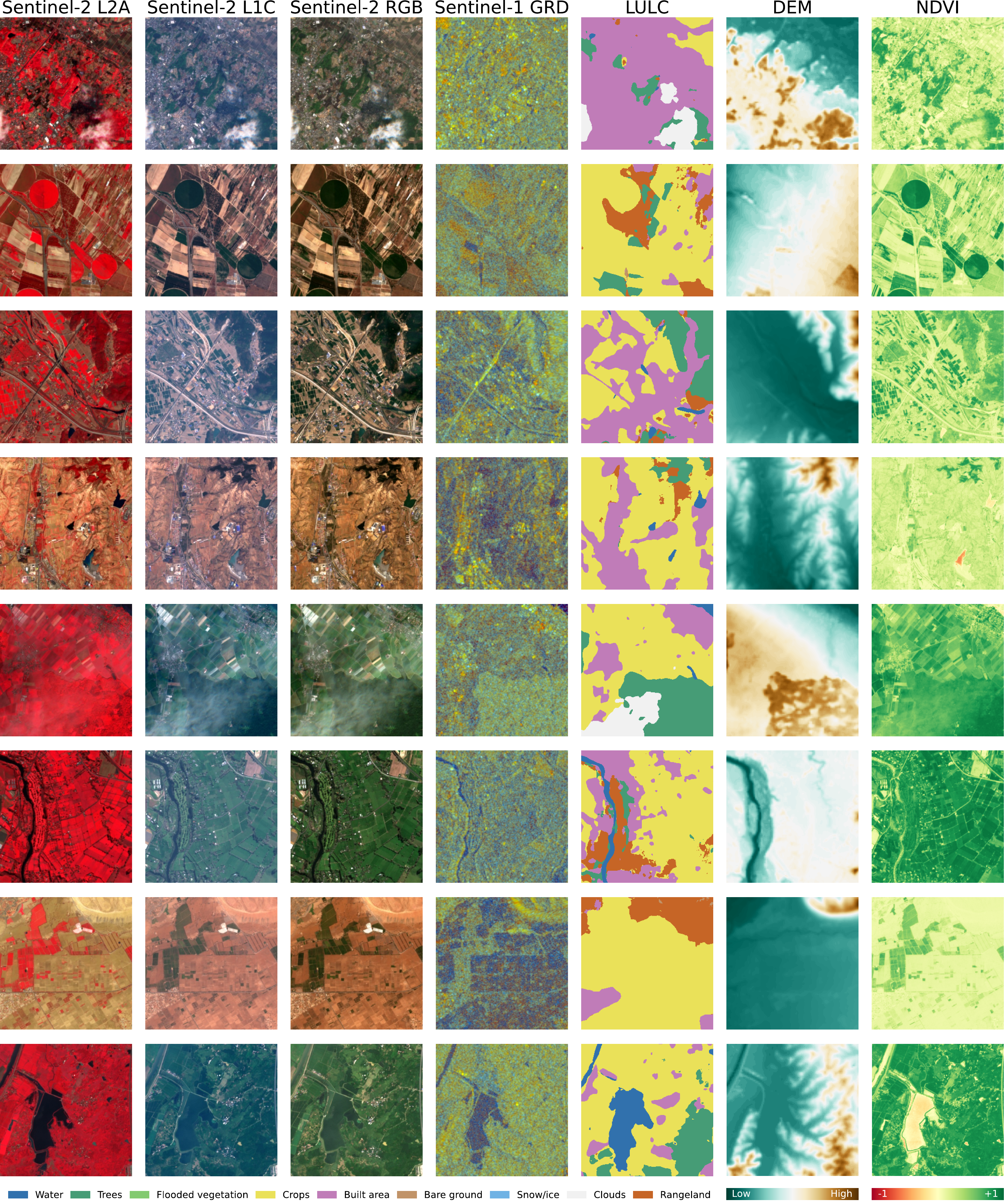}
    \captionof{figure}{Additional examples from the SSL4EO-S12 subset with seven spatiotemporal aligned modalities. Sentinel-2 L2A uses IRRG pseudo-coloring, and Sentinel-1 RTC is visualized in dB scale as VH-VV-VV/VH. Copernicus DEM is scaled based on the image value range with an additional 10\,meter buffer.}
    \label{fig:ssl4eo_samples}
\end{figure*}

On the following pages, we present additional examples from both data sources in Figure~\ref{fig:majortom_samples} and~\ref{fig:ssl4eo_samples}. All examples are taken from each subset's first validation tar file, and we manually select representative samples to showcase the full range of landscapes. 
The focus on urban areas in SSL4EO-S12 is visible, which benefits the pre-training compared to homogeneous sampling.
Because of the additional scaling for S-2 RGB, images with few clouds are not very dark, while the clouds keep their structure and are not flattened to a single white value as it is the case with clipping to a max value.
Notice the alignment of the cloud and ice classes in the LULC data due to the augmentation using the SEnSeI~v2 cloud and ice masks. We only annotate thick clouds as the land surface is still visible below thin cloud layers.

